# Unified Kernel-Segregated Transpose Convolution Operation

Vijay Srinivas Tida
University of Louisiana at Lafayette
vijay-srinivas.tida1@louisiana.edu

Md Imran Hossen
University of Louisiana at Lafayette
md-imran.hossen1@louisiana.edu

Liqun Shan
University of Louisiana at Lafayette
liqun.shan1@louisiana.edu

Sai Venkatesh Chilukoti
University of Louisiana at Lafayette
sai-venkatesh.chilukoti1@louisiana.edu

Sonya Hsu
University of Louisiana at Lafayette
hsiu-yueh.hsu@louisiana.edu

Xiali Hei
University of Louisiana at Lafayette
xiali.hei@louisiana.edu

## Abstract

*The optimization of the transpose convolution layer for deep learning applications is achieved with the kernel segregation mechanism. However, kernel segregation has disadvantages, such as computing extra elements to obtain the output feature map with odd dimensions while launching a thread. To mitigate this problem, we introduce a unified kernel segregation approach that limits the usage of memory and computational resources by employing one unified kernel to execute four sub-kernels. The findings reveal that the suggested approach achieves an average computational speedup of $2.03\times$ ($3.89\times$) when tested on specific datasets with an RTX 2070 GPU (Intel Xeon CPU). The ablation study shows an average computational speedup of $3.5\times$ when evaluating the transpose convolution layers from well-known Generative Adversarial Networks (GANs). The implementation of the proposed method for the transpose convolution layers in the EB-GAN model demonstrates significant memory savings of up to 35 MB.*

## 1. Introduction

Traditional convolution and transpose convolution operations involve accumulation processes (multiplications and additions), which are especially important in image processing techniques for deep learning applications Gaudenz and Nico (5/26/2023 accessed). Applying a convolution operation to the input feature map decreases the size of the output feature map. Conversely, using a transpose convolution operation on the input feature map increases the size of the output feature map. To put it differently, the transpose convolution operation aids in transforming low-resolution images into high-resolution images. This study employs transpose convolution with a stride of one to generate the output feature map. Fig. 1 illustrates the basic convolution and transpose convolution processes when executed on an input feature map of dimensions $4 \times 4$. Deep learning models face checkboard problems when using the standard transpose convolution operation with stride one Zhou (4/8/2023 accessed). The checkerboard issue occurs due to uneven accumulation of values at certain points in the output feature map. Thus, to mitigate the checkerboard problem, researchers devised a method that integrates an upsampling layer with a convolution operation.

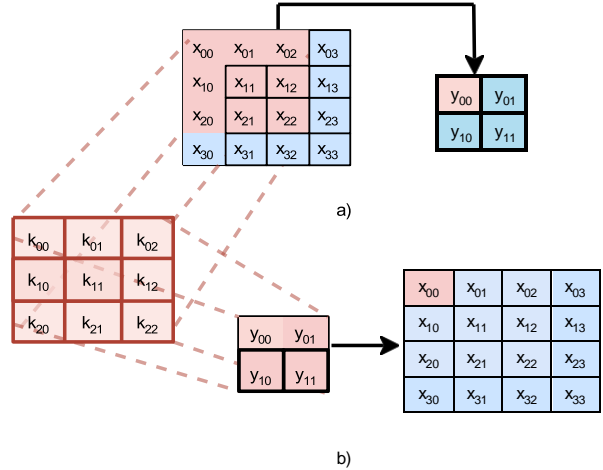

Figure 1: **a)** Conventional convolution and **b)** Transpose convolution.

The transpose convolution with a stride of two has emerged as the most commonly used component in GAN generators for image creation. In this version of transpose convolution, the upsampling layer is followed by the convolution layer. The upsampling layer is created by inserting zeros between each row and column. This approach is known as the bed-of-nails approach, as stated in Yazdanbakhsh, Samadi, et al. (2018). Therefore, an input feature map with dimensions $N \times N$ is transformed into $(2N-1) \times (2N-1)$ after the up-sampling layer. Thus, applying the normal convolution operation with a kernel size of $n \times n$ with stride one leads to the output feature map of size $(2N-n) \times (2N-n)$. Fig. 2 gives an overview of the basic transpose convolution operation with the input feature map of 4×4 and a kernel of 3×3. The redundant zeros in the upsampled feature map lead to excessive data transfers, memory bottlenecks, and inefficient use of computing resources. The unified kernel mechanism introduced in this paper addresses the aforementioned issues more effectively at the algorithmic level than previous methods. As a result, the proposed approach eliminates the need for developing any new hardware accelerators.

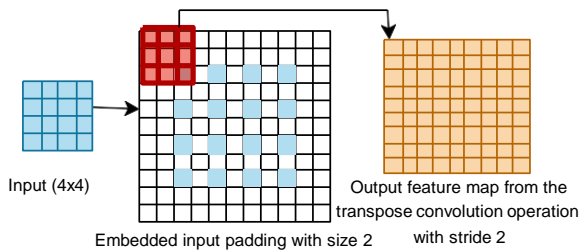

Figure 2: **Transpose convolution operation with padding factor of 2.**

Researchers face real-time challenges in accessing the designed hardware accelerators for efficient implementation of the transpose convolution layer V. HUYNH (U.S. Patent WO/2021/061566, April. 2021), Yazdanbakhsh, Brzozowski, et al. (2018), and Yazdanbakhsh, Samadi, et al. (2018). The work presented in V. S. Tida, Chilukoti, Hsu, et al. (2023a) implements an efficient transpose convolution layer using a kernel segregation mechanism at the algorithmic level. However, the main drawback of the kernel segregation mechanism is that it produces extra elements if the output feature size is an odd number of dimensions. The main reason for that is grouping four sub-kernels in a single thread. This paper addresses this issue by using the unified kernel segregation mechanism for computing the transpose convolution layer. During runtime, the process selects a sub-kernel rather than pre-defining the four sub-kernels as done previously. This new implementation achieves significant memory savings by eliminating unnecessary computations in specific scenarios. The significant contributions of this paper are as follows:

a. We introduce an optimized transpose convolution algorithm that employs a unified kernel segregation approach to minimize computational overhead and memory usage without the need for specialized hardware accelerators.

b. We analyze the speed up of computation time and memory consumption of our proposed approach using multiple datasets. The results show substantial memory savings, especially for the output feature of an odd number of dimensions.

c. We investigate the performance of the proposed optimization using transpose convolution layers from popular GANs. Our experimental results indicate a significant improvement of more than 3× in computation speed compared to the naive approach.

The rest of the paper is organized as follows: Section 2 explains the background and literature review, while Section 3 illustrates the proposed unified kernel segregation mechanism to implement transpose convolution. Then, Section 4 interprets the results, and Section 5 addresses limitations and future research directions. Finally, we will conclude the paper in Section 6.

## 2. Background and Literature Review

The evolution of deep learning models has shown a significant improvement in the healthcare industry Chilukoti et al. (2024), Internet of Things (IoT) devices V. S. S. Tida et al. (2022), image generation tasks Li et al. (2022), natural language processing applications V. S. Tida and Hsu (2022) and V. S. Tida, Hsu, and Hei (2023), etc. Recently, maintaining privacy and security for these models has become more important in protecting individual safety and devices from unauthorized access V. S. Tida, Chilukoti, Hsu, et al. (2023b) and V. S. S. Tida et al. (2022). Consequently, the computation time becomes a significant overhead, primarily due to the increased number of accumulation operations. Actually, a majority of these values may not significantly enhance the model's performance. The primary objective of the optimization process is to avoid redundant computations, enabling faster model training with minimal accuracy loss. Thompson et al. (2020). Thus, it also helps to utilize the hardware resources efficiently according to the user requirements.

The optimization of deep learning models can be grouped into two main classes. The first category is known as exact optimization, where the computed output remains unchanged but minimizes unnecessary overhead or makes efficient use of hardware resources V. S. Tida et al. (2022) Anderson et al. (2020). The unified kernel segregation mechanism for optimizing the transpose convolution layer computation belongs to this class. The second class, namely approximate optimization, mainly focuses on reducing the computation load by sacrificing the performance of the models M. S. Kim et al. (2021). The primary objective is to develop optimized models capable of running on handheld devices like mobile phones, thereby facilitating the widespread dissemination of the application.

Optimization techniques for deep learning models can be applied in three stages: data preprocessing,

hyperparameter tuning, and model selection Menghani (2023). Based on the user's requirements, the models are designed by considering different stages individually or in combination. However, when approximation methods are considered, they might degrade the model's performance. The work presented here is similar to that of V. S. Tida, Chilukoti, Hsu, et al. (2023a) and Vasudevan et al. (2017) where the optimization is performed at the data pre-processing stage. The rearrangement of kernel elements eliminates unnecessary computations before implementing the transpose convolution. The proposed optimized method significantly reduces computational load and memory requirements in the training and inference stages without performance degradation.

### 2.1. Existing works for optimizing transpose convolution

Initially, researchers focused on designing hardware accelerators to implement efficient transpose convolution layers that involve up-sampling and convolution layers. It is crucial to enhance the efficiency of transpose convolution with the current hardware resources due to their limited availability. The first hardware accelerator introduced by Yazdanbakhsh, Samadi, et al. (2018) employs a method of rearranging the output and filter rows. This process helps to avoid almost half of the ineffective computations for transpose convolution layers. The efficiency of the proposed accelerator is evaluated using well-known GANs as a possible benchmark for measuring the computation time. The suggested hardware accelerator requires the integration of Single Instruction Multiple Data (SIMD) and Multiple Instruction Multiple Data (MIMD) architectures. The findings indicated that the proposed accelerator achieved an average speedup of $3.6\times$ for generative deep learning models in comparison to the Eyeriss hardware accelerator. Later, the same group of researchers utilized Field Programmable Gate Arrays (FPGAs) by further developing the rearrangement technique described in Yazdanbakhsh, Brzozowski, et al. (2018). The results demonstrated a performance increase of $2.2\times$ over the optimized conventional accelerator and $2.6\times$ over the Titan X GPU.

Recently, Amazon Technologies Inc. filed a patent V. HUYNH (U.S. Patent WO/2021/061566, April. 2021), showing the efficient implementation of transpose convolution using systolic arrays. The main disadvantage of these approaches is that some devices might need to use the upsampled layer. These upsample layers are generated with the help of the bed-of-nails approach applied to the input feature map. The extra load is also introduced during the backward propagation phase when training the deep neural network. Additionally, these methods require specialized hardware to efficiently perform transpose convolution, which may not be easily accessible or available. Moreover, it may not be practical to have distinct hardware dedicated to a single application.

The optimized algorithmic approach to compute transpose convolution using a segregation mechanism was proposed in V. S. Tida, Chilukoti, Hsu, et al. (2023a). However, the implementation is concerned with generating extra elements during the computation process if the output feature is of an odd number of dimensions, especially for GPU applications. The work presented here addresses the above-mentioned problem by selecting a sub-kernel during the runtime to prevent extra elements and save memory resources.

## 3. Methodology

### 3.1. Kernel segregation mechanism

Transpose convolution implementation is equivalent to four internal convolution operations applied to the standard image placed at different offsets without using an up-sampling technique. These offsets can be regarded as the starting point of the input image to apply the convolution using the achieved sub-kernels from the original kernel. Thus, segregating the original kernel into four sub-kernels can be analyzed using Fig. 3. Here, the input feature map is upsampled using a bed-of-nails approach, and then a computation pattern for four general cases is observed by ignoring the padding elements. The green squares indicate that the multiplication operations are necessary at the respective locations. The red squares help identify the ineffective multiplications involved at the positions where the kernel elements need to be rearranged accordingly. Thus, the input feature map does not require an up-sampling process to apply transpose convolution operation with these modifications on the original kernel. Finally, the original kernel is segregated into four sub-kernels by considering the elements positioned at successive rows and columns.

In Fig. 3, we ignore the padding elements and analyze the computation pattern by examining only the upsampled input feature map. When a kernel of size $5 \times 5$ is placed on the feature map, it shows only nine effective multiplications in the first case, as seen in Fig. 3a. Similarly, when the kernel is transferred to the adjacent side, only six multiplications are required, and others must be discarded, as shown in Fig.

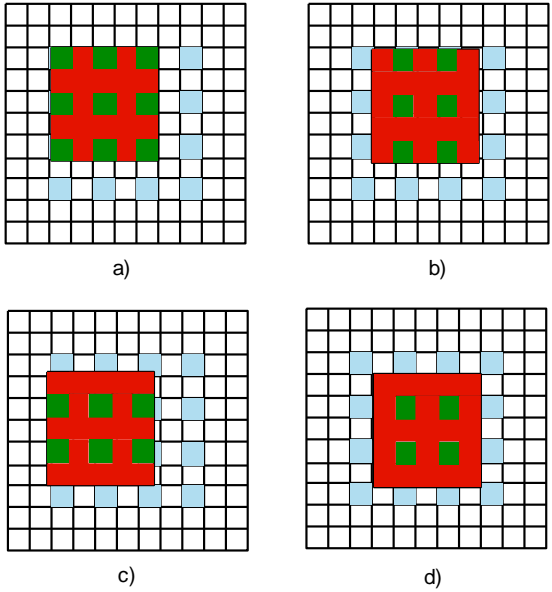

Figure 3: **Transpose convolution operation. a), b), c), and d) show the computation pattern throughout the input feature map.**

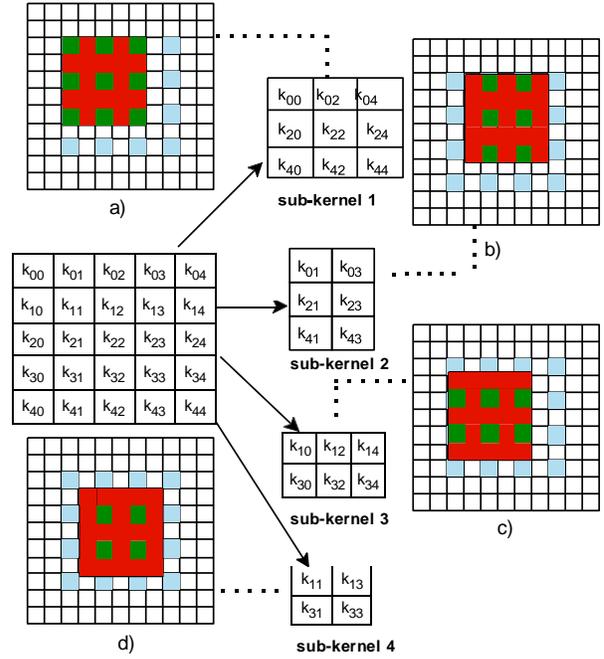

Figure 4: **Kernel segregation mechanism for $5 \times 5$ kernel.**

3b. This process of computation continues repeatedly until the completion of this row. However, different combinations of multiplications can be observed in the next row. Thus, in Fig. 3c, only six multiplications are needed to compute the particular element in the output feature map. Note here that the active element pattern for the kernel differs from the above case. Similarly, as in Fig. 3d, only four multiplications are needed for computation to get the output element. Thus, the same computation pattern continues for the entire row to produce elements in the output feature map. The four computation patterns can be seen throughout the feature map such that the starting index depends on the rows and columns at the relevant location in the input feature map. Thus, the proposed approach uses 25 multiplications efficiently to produce four output elements by performing four internal convolutions, compared to only one from the conventional approach.

### 3.2. Kernel segregation mechanism for $5 \times 5$ kernel

Based on the effective computation patterns, the single large kernel for the transpose convolution will be converted into four sub-kernels. The process is outlined in Fig. 4 considering the kernel's size of $5 \times 5$. The sub-kernels are formed by considering the elements from consecutive rows and columns at the specific elements of the locations, which are indicated by green color. Thus, nine and six elements form the first and second sub-kernels. Similarly, the third and fourth sub-kernels are created with six and four elements based on their active positions. The odd-ordered kernel is examined here because the sizes obtained for four sub-kernel sizes will vary. Also, the same process applies to even ordered kernels without restrictions to create four sub-kernels of the same size. The problem with the previous proposed approach in implementing the transpose convolution is that the accumulations from four sub-kernels are calculated sequentially using a single thread at run time V. S. Tida, Chilukoti, Hsu, et al. (2023a). Thus, it results in extra memory usage if the output feature map has odd dimensions. The unified kernel segregation algorithm uses the selection process of the exact sub-kernel for each thread during the kernel launch. Thus, this approach can eliminate ineffective computations and memory wastage to a greater extent.

### 3.3. Conventional and unified kernel segregation algorithms for implementing the transpose convolution

The pseudocode for implementing the conventional transpose convolution is shown in Algorithm 1. Initially, the input feature map of size $N \times N$ and the kernel of size $n \times n$ are considered. After applying the transpose convolution operation, the variable $out$ gives the output feature map. The input feature map is upsampled using the bed-of-nails approach by filling zeros alternately row-wise and column-wise. Thus, the upsampled feature map of size $(2N-1) \times (2N-1)$ is obtained from the actual size of $N \times N$. The symbol $\circledast$ represents the convolution operation applied on the

**Algorithm 1** Conventional Transpose Convolution algorithm

**Input:** Input feature map I$\{x_{00}, ..., x_{N-1,N-1}\}$ of size $N \times N$, original kernel $K$ of size $n \times n$
**Initialize** $K$ randomly
Upsampled input feature map $U$ of size $(2N-1) \times (2N-1)$ initialized with zeros
**for** $i = 0; i < N; i++$ **do**
  **for** $j = 0; j < N; j++$ **do**
    $U_{2i,2j} = I_{i,j}$
  **end**
**end**
**for** $i = 0; i < (2N-n); i++$ **do**
  **for** $j = 0; j < (2N-n); j++$ **do**
    $out_{i,j} = U_{i:i+n, j:j+n} \circledast K_{n,n}$
  **end**
**end**
**Output:** $out$ represents the output feature map upsampled input feature map to get the output feature map of size $(2N-n) \times (2N-n)$.

**Algorithm 2** Unified Kernel Segregated Transpose Convolution algorithm

**Input:** Input feature map I$\{x_{00}, ..., x_{N-1,N-1}\}$ of size $N \times N$, segregated sub-kernels $k_{00}, k_{01}, k_{10}$ and $k_{11}$ of sizes $\lceil n/2 \rceil \times \lceil n/2 \rceil$, $\lceil n/2 \rceil \times \lfloor n/2 \rfloor$, $\lfloor n/2 \rfloor \times \lceil n/2 \rceil$ and $\lfloor n/2 \rfloor \times \lfloor n/2 \rfloor$ respectively, where $n \times n$ is the size of the original kernel $K$.
**Initialize** $k_{00}, k_{01}, k_{10}$ and $k_{11}$ randomly

**for** $i = 0; i < (2N-n); i++$ **do**
  **for** $j = 0; j < (2N-n); j++$ **do**
    $r \leftarrow i\%2; s \leftarrow j\%2$
    $R \leftarrow row\_size(k_{r,s})$
    $C \leftarrow column\_size(k_{r,s})$
    $out_{i,j} = I_{\lceil \frac{i}{2} \rceil:\lceil \frac{i}{2} \rceil+R, \lceil \frac{j}{2} \rceil:\lceil \frac{j}{2} \rceil+C} \circledast k^{rs}_{R,C}$
  **end**
**end**
**Output:** $out$ represents the output feature map

Similarly, the proposed unified kernel segregation mechanism produces the same output feature map without upsampling the input data. The pseudocode for the segregated transpose convolution of the unified kernel segregated transpose convolution is shown in Algorithm 2. The inputs to the unified kernel segregated transpose are the input feature map and the four segregated sub-kernels. The four segregated sub-kernels are formed based on the computation pattern by analyzing every other row and column during convolution operations. Four sub-kernel sizes will be formulated based on alternatively selecting elements along the rows and columns from the original kernel. The appropriate sub-kernel is chosen during runtime using variables $r$ and $s$ ($r = i\%2$ and $s = j\%2$).

Based on these values, the row and column sizes are obtained, and then the convolution operation is applied to the modified input feature map with the specified sub-kernel at runtime. The output feature map obtained will be the same as the conventional transpose convolution implementation. Here, the padding of the input feature map is ignored for both cases. The padding factor for the optimized algorithm is different from the conventional approach and will be illustrated in the next section.

### 3.4. Optimization of the transpose convolution operation using segregated kernels

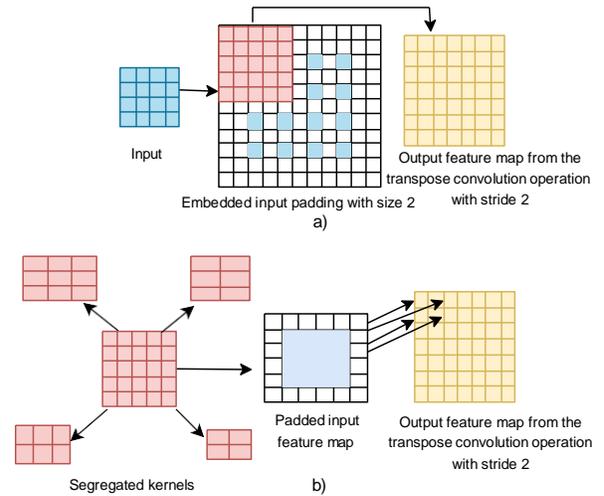

Figure 5: **Comparison of the a) conventional transpose convolution with the b) proposed optimized technique.**

The conventional and proposed optimization procedures for implementing transpose convolution are depicted in Fig. 5. The proposed implementation has four segregated kernels instead of a single kernel, and four internal convolutions are applied on the input feature map. However, the proposed algorithm reduces the padding factor to almost half the original. The reduction in the padding factor helps prevent unnecessary output elements from the convolutions during accumulation operations. In Fig. 5, we consider an input feature map of $4 \times 4$ and a kernel size of $5 \times 5$ in both cases. The conventional transpose convolutional algorithm needs a padding factor of two for the obtained upsampled feature map, which can be seen in Fig. 5a. In contrast, the proposed method does not need the upsampled input feature map. Instead, kernels are segregated to form four sub-kernels based on the computation pattern shown in Fig. 5b. In addition, the padding factor for the input feature using four segregated kernels differs from the original. For example, if the initial padding factor is $P$, the new padding factor becomes $\lfloor P/2 \rfloor$. Additionally, if the original padding factor is odd, then the sub-kernel order changes to $k_{11}$, $k_{10}$, $k_{01}$, and $k_{00}$ from $k_{00}$, $k_{01}$, $k_{10}$, and $k_{11}$.

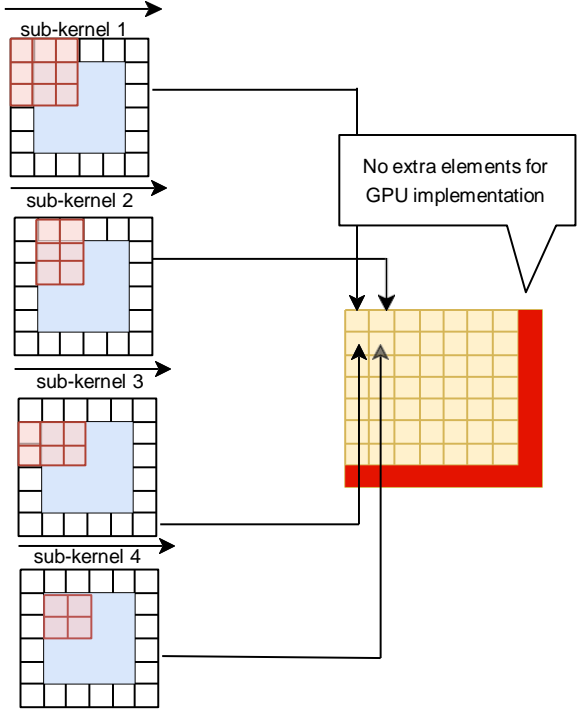

Figure 6: **Optimized transpose convolution implementation using four sub-kernels with input size** $4 \times 4$ **and padding with factor 1 (but padding factor is 2 for the original case).**

The proposed optimized technique using a unified kernel segregation algorithm applied to the input feature map can be observed in Fig. 6. The transpose convolution operation is performed by selecting the sub-kernel during the runtime on an input feature of size $4 \times 4$. In addition, the padding factor will be reduced from two to one because the input feature map is used directly to obtain a similar output feature map. Finally, the four sub-kernels perform the accumulation operations individually instead of grouping them based on the exact location of the output feature map. In an ideal case, the proposed method should show a four-times speedup in computation compared to the conventional method. However, the proposed method has an additional computation burden because of appropriate sub-kernel identification during runtime. The proposed method must also find the exact locations from the output feature map. These extra tasks might reduce the computation speed compared to the ideal scenario.

The prior version V. S. Tida, Chilukoti, Hsu, et al. (2023a) has a problem of generating extra output elements when the output feature map is of odd dimensions. The proposed implementation addresses these limitations by running only one convolution operation on a single thread. The formulas for calculating the four output feature values with the help of the proposed optimization process can be expressed in Equations 1- 4.

$$out[x][y] = \sum_{u=0}^{a} \sum_{v=0}^{a} in[i+u][j+v] \cdot k_{00}[u][v],$$
$$where\ x = 2 \cdot i, y = 2 \cdot j \quad (1)$$

$$out[x][y] = \sum_{u=0}^{a} \sum_{v=0}^{b} in[i+u][(j+1)+v] \cdot k_{01}[u][v],$$
$$where\ x = 2 \cdot i, y = 2 \cdot j + 1 \quad (2)$$

$$out[x][y] = \sum_{u=0}^{b} \sum_{v=0}^{a} in[(i+1)+u][j+v] \cdot k_{10}[u][v],$$
$$where\ x = 2 \cdot i + 1, y = 2 \cdot j \quad (3)$$

$$out[x][y] = \sum_{u=0}^{b} \sum_{v=0}^{b} in[(i+1)+u][(j+1)+v] \cdot k_{11}[u][v],$$
$$where\ x = 2 \cdot i + 1, y = 2 \cdot j + 1 \quad (4)$$

where $a = \lceil n/2 \rceil - 1$, $b = \lfloor n/2 \rfloor - 1$, $out[x][y]$ represents the output feature map located in the $x^{th}$ row and $y^{th}$ column; $in[i][j]$ represents the input feature map in the $i^{th}$ row and $j^{th}$ column; $k_{00}[u][v]$, $k_{01}[u][v]$, $k_{10}[u][v]$ and $k_{11}[u][v]$ represent the subkernels $k_{00}$, $k_{01}$, $k_{10}$ and $k_{11}$ obtained after the segregation mechanism and their locations in the $u^{th}$ row and $v^{th}$ column. The sizes of the four acquired subkernels will be $\lceil n/2 \rceil \times \lceil n/2 \rceil$, $\lceil n/2 \rceil \times \lfloor n/2 \rfloor$, $\lfloor n/2 \rfloor \times \lceil n/2 \rceil$, and $\lfloor n/2 \rfloor \times \lfloor n/2 \rfloor$. Here, the size of the input feature map remains the same without upsampled values. However, applying the proposed optimization method reduces the padding factor to half of the original implementation.

## 4. Results

### 4.1. Datasets and the evaluation procedure

The images from the Flower dataset Mamaev (4/8/2022 accessed), MSCOCO 2017 "Common Objects in Context Dataset" (8/26/2022 accessed), and PASCAL

Table 1: **Considered datasets characteristics**

| Dataset | | Number of samples |
|---|---|---|
| Flowers | Sunflower | 734 |
| | Tulip | 984 |
| | Daisy | 769 |
| | Rose | 784 |
| | Dandelion | 1,052 |
| MSCOCO (2017) | (used 10% of total samples) | 11,828 |
| PASCAL (2017) | Classification | 17,125 |
| | Segmentation | 2,913 |

VOC 2012 "Visual Object Classes Challenge 2012" (8/26/2022 accessed) datasets are considered to evaluate the importance of the proposed unified optimization method. The metrics, such as speed up in computation time and memory savings, help to analyze the proposed optimization method over the conventional implementation. The computation time for all the cases is recorded in seconds. The memory savings are obtained by calculating the difference between the memory requirement made by the original and the proposed implementations for the transpose convolution layers involved. Detailed information on the data sets can be shown in Table 1. The images are converted to a standard format of $224 \times 224 \times 3$, and transpose convolution operations are performed using conventional and proposed unified approaches. The evaluation platform uses the C++ and CUDA C programming languages for CPU and GPU.

### 4.2. Analysis of the proposed unified kernel segregation algorithm

We consider computation time and memory requirements metrics to analyze the benefits of the proposed unified approach over the conventional method. Tables 2 and 3 provide information on the speed-up and memory savings obtained using a unified method over the implementation of naive transpose convolution. The proposed algorithms are evaluated using the Intel Xeon CPU and NVIDIA GeForce RTX 2070 GPU devices. The results are obtained with the help of the GCC 9.4 and CUDA 10.2 versions using the CPU and GPU, respectively. The advantages of the proposed unified method are analyzed by varying the kernel size from $3 \times 3$ to $5 \times 5$ by converting all the image samples to the standard size of $224 \times 224 \times 3$. The computation time and memory savings are reported using three datasets, and it was observed that the memory savings are similar for all the cases without being dependent on the structure of the output feature map. The proposed method showed an average speedup of more than $2\times$ in computation time compared to the conventional method using the GPU implementation.

The main reason for the limited speedup compared to the previously proposed approach for GPU is because of extra memory loads, fewer kernels, and complex computation instructions added for each thread. The added burden on scheduling threads' launch for small computation jobs that produce only one output feature map. However, there is an improvement in speedup when kernels increase to have more output feature maps, which can be explained in the ablation study.

On the other hand, a faster speedup than above $3.9\times$ is achieved for CPU implementation. Even though the output feature map obtained from the GPU does not have extra output elements, the limited computation capability showed a reduced speedup compared to the CPU method. The speedup obtained from $4 \times 4$ is higher than other kernels, possibly due to the fact that the memory bandwidth is well-fitted to transfer the data elements during the computation process. The memory savings for the implementation in Table 3 are still the same as in Table 2 as the input feature map is of the same size.

### 4.3. Ablation study

For the analysis of the ablation study, the convolution layers transpose from the popular GAN models are considered Yazdanbakhsh, Samadi, et al. (2018). The speedup in computation time and the memory savings obtained from the proposed unified approach compared to the conventional implementation can be seen in Table 4. The number of floating-point operation reductions remains the same as the work presented in V. S. Tida, Chilukoti, Hsu, et al. (2023a). The computation time and memory savings are calculated only for the forward propagation stage for the transpose convolution layers using one input image sample. The results show a speedup of $3.06\times$ ($4.21\times$) for the DC-GAN/DiscoGAN T. Kim et al. (2017) and Radford et al. (2015) models for the GPU (CPU) version. The memory savings obtained from the transpose convolution layers for DC-GAN/DiscoGAN models is around 4,787,712 bytes.

Similarly, Art-GAN Tan et al. (2017), GP-GAN Wu et al. (2019), and EB-GAN Zhao et al. (2016) show a substantial speedup of more than $2.67\times$ in computation time for GPU and $4\times$ for CPU devices. The memory savings obtained from these models for the transpose convolution layers are also clearly observed in Table 4. These models also achieved an average speedup of $2.95\times$ ($4.2\times$) implemented using a GPU (CPU) device. The proposed unified algorithm addresses memory inefficiency when the output feature map has odd dimensions. The limitation of achieving this capability is that it sacrifices speed-up due to additional computational load for GPU applications. However, the computation speed for the selected GANs increases based on the number of floating point operations involved.

Table 2: **Speedup for GPU and CPU versions and memory savings obtained for the Flower dataset for the conventional (Conv) and the Proposed (Prop) approaches**

| Data group | Kernel | Conv (GPU) | Conv (CPU) | Prop (GPU) | Prop (CPU) | Speedup (GPU) | Speedup (CPU) | Memory savings (MegaBytes) |
|---|---|---|---|---|---|---|---|---|
| Daisy | 5 × 5 × 3 | 3.6233 | 61.388 | 1.780 | 15.143 | 2.035 | 4.054 | 1.8279 |
| | 4 × 4 × 3 | 2.715 | 38.978 | 1.0997 | 10.016 | 2.47 | 3.892 | 1.8279 |
| | 3 × 3 × 3 | 1.7454 | 22.491 | 1.11 | 5.899 | 1.572 | 3.812 | 1.8279 |
| Dandelion | 5 × 5 × 3 | 5.073 | 84.122 | 2.482 | 20.662 | 2.044 | 4.071 | 1.8279 |
| | 4 × 4 × 3 | 3.7712 | 53.573 | 1.546 | 13.618 | 2.44 | 3.934 | 1.8279 |
| | 3 × 3 × 3 | 2.6008 | 30.978 | 1.5508 | 8.051 | 1.677 | 3.848 | 1.8279 |
| Rose | 5 × 5 × 3 | 3.5505 | 63.06 | 1.8 | 15.476 | 1.972 | 4.075 | 1.8279 |
| | 4 × 4 × 3 | 2.736 | 39.945 | 1.116 | 10.166 | 2.45 | 3.929 | 1.8279 |
| | 3 × 3 × 3 | 1.9034 | 23.081 | 1.138 | 6.01 | 1.673 | 3.840 | 1.8279 |
| Sunflower | 5 × 5 × 3 | 3.3974 | 58.809 | 1.687 | 14.408 | 2.014 | 4.082 | 1.8279 |
| | 4 × 4 × 3 | 2.564 | 37.438 | 1.0253 | 9.534 | 2.501 | 3.927 | 1.8279 |
| | 3 × 3 × 3 | 1.6225 | 21.442 | 1.0524 | 5.818 | 1.541 | 3.686 | 1.8279 |
| Tulip | 5 × 5 × 3 | 4.5116 | 79.113 | 2.282 | 19.356 | 1.977 | 4.087 | 1.8279 |
| | 4 × 4 × 3 | 3.5148 | 49.918 | 1.437 | 12.822 | 2.446 | 3.893 | 1.8279 |
| | 3 × 3 × 3 | 2.2988 | 28.734 | 1.4295 | 7.499 | 1.608 | 3.832 | 1.8279 |

Table 3: **Speedup for GPU and CPU versions for MSCOCO and PASCAL datasets using the conventional (Conv) and the proposed (Prop) approaches**

| Dataset | Kernel | Conv (GPU) | Conv (CPU) | Prop (GPU) | Prop (CPU) | Speedup (GPU) | Speedup (CPU) |
|---|---|---|---|---|---|---|---|
| MSCOCO 2017 | 5 × 5 × 3 | 64.150 | 914.852 | 29.9012 | 228.449 | 2.1453 | 4.005 |
| | 4 × 4 × 3 | 43.859 | 579.72 | 18.321 | 149.448 | 2.394 | 3.879 |
| | 3 × 3 × 3 | 29.0158 | 327.78 | 18.877 | 88.625 | 1.537 | 3.698 |
| PASCAL VOC 2012 (Classification) | 5 × 5 × 3 | 90.643 | 1302.201 | 44.4184 | 331.762 | 2.041 | 3.925 |
| | 4 × 4 × 3 | 63.4066 | 822.349 | 26.546 | 215.781 | 2.38 | 3.811 |
| | 3 × 3 × 3 | 42.291 | 474.865 | 27.423 | 127.36 | 1.54 | 3.728 |
| PASCAL VOC 2012 (Detection) | 5 × 5 × 3 | 13.665 | 223.304 | 7.1315 | 55.919 | 1.916 | 3.993 |
| | 4 × 4 × 3 | 10.619 | 141.742 | 4.395 | 36.887 | 2.416 | 3.8426 |
| | 3 × 3 × 3 | 7.0369 | 81.742 | 4.5286 | 21.834 | 1.554 | 3.744 |

## 5. Discussion

The proposed unified approach has limitations as it requires additional instructions and an array to store the four sub-kernel sizes primarily used in GPU implementation. The array takes a maximum of 32 bytes and might not be a significant overhead to implement, as models generally involve multiple transpose convolution layers. These steps will help to ensure the exact sub-kernel selection during the runtime while performing computations. If the user demands less memory usage for the generation of output feature maps, the unified kernel segregation mechanism will be the best choice to achieve the goal. However, reducing the extra workload and avoiding memory usage for storing the four sub-kernel sizes can be treated as future research direction to achieve maximum performance.

The transpose convolution using the matrix multiplication method can utilize the proposed mechanism Wang et al. (2019). However, the major problem with matrix multiplication is identifying the position of the output elements involved when multiple threads run in parallel. This process will result in four subarrays for the output feature map by every other row and column. Thus, this method requires an extra step to rearrange the elements to get the same output elements from the original implementation. This process demands more memory, which might be equivalent to double the size of the output feature map, and requires more time to obtain the exact output feature map.

In dilated convolution, the kernels are upsampled using a bed-of-nails approach, and then the convolution operation is applied Yu and Koltun (2015). The same computation pattern approach can be applied by utilizing the segregated input feature maps, and kernels remain the same without any modifications. Thus, this process will save memory resources and eliminate unnecessary computations involved in the naive approach for dilated convolution. In addition, the matrix multiplication method for dilated convolution can follow the proposed optimization process and be seen as a potential future research direction.

Table 4: **Speedup for GPU and CPU versions and Memory savings obtained from transpose convolution layers for popular GAN models using the unified approach**

| Model | # | Input size | Kernel size | Conv (GPU) | Prop (GPU) | Conv (CPU) | Prop (CPU) | Memory savings (Bytes) |
|---|---|---|---|---|---|---|---|---|
| DC-GAN/ DiscoGAN | 2 | 4 × 4 × 1024 | 4 × 4 × 1024 × 512 | 0.046753 | 0.01530 | 3.023 | 0.727 | 495,616 |
| | 3 | 8 × 8 × 512 | 4 × 4 × 512 × 256 | 0.046085 | 0.01488 | 3.101 | 0.7102 | 739,328 |
| | 4 | 16 × 16 × 256 | 4 × 4 × 256 × 128 | 0.043747 | 0.01406 | 2.90 | 0.704 | 1,254,400 |
| | 5 | 32 × 32 × 128 | 4 × 4 × 128 × 3 | 0.003049 | 0.001391 | 0.1363 | 0.0341 | 2,298,368 |
| | | Total | | 0.139639 | 0.045631 | 9.1603 | 2.1753 | |
| | | Speedup/ Memory saved | | | 3.0601 | | 4.211 | 4,787,712 |
| ArtGAN | 2 | 4 × 4 × 512 | 4 × 4 × 512 × 256 | 0.011886 | 0.004580 | 0.7114 | 0.1890 | 4,247,808 |
| | 3 | 8 × 8 × 256 | 4 × 4 × 256 × 128 | 0.011726 | 0.004160 | 0.7219 | 0.1780 | 369,664 |
| | 4 | 16 × 16 × 128 | 4 × 4 × 246 × 128 | 0.021727 | 0.00735 | 1.3879 | 0.330 | 627,200 |
| | 6 | 32 × 32 × 128 | 4 × 4 × 128 × 3 | 0.00142 | 0.001432 | 0.0359 | 0.0075 | 67,200 |
| | | Total | | 0.046921 | 0.017522 | 2.8571 | 0.7034 | |
| | | Speedup/ Memory saved | | | 2.67 | | 4.06184 | 1,871,872 |
| GP-GAN | 2 | 4 × 4 × 512 | 4 × 4 × 512 × 256 | 0.011847 | 0.00468 | 0.7114 | 0.194 | 247,808 |
| | 3 | 8 × 8 × 256 | 4 × 4 × 256 × 128 | 0.01171 | 0.004050 | 0.7219 | 0.1692 | 369,664 |
| | 4 | 16 × 16 × 128 | 4 × 4 × 128 × 64 | 0.01167 | 0.004102 | 0.6995 | 0.1682 | 627,200 |
| | 5 | 32 × 32 × 64 | 4 × 4 × 64 × 3 | 0.001574 | 0.000782 | 0.0659 | 0.016 | 1,149,184 |
| | | Total | | 0.036801 | 0.0138614 | 2.1987 | 0.5474 | |
| | | Speedup/ Memory saved | | | 2.703 | | 4.0166 | 2,393,856 |
| EB-GAN | 2 | 4 × 4 × 2048 | 4 × 4 × 2048 × 1024 | 0.188821 | 0.05961 | 16.0994 | 3.692 | 991,232 |
| | 3 | 8 × 8 × 1024 | 4 × 4 × 1024 × 512 | 0.176702 | 0.0543 | 14.193 | 2.93 | 1,478,656 |
| | 4 | 16 × 16 × 512 | 4 × 4 × 512 × 256 | 0.172839 | 0.05321 | 16.587 | 2.83 | 2,508,800 |
| | 5 | 32 × 32 × 256 | 4 × 4 × 256 × 128 | 0.172694 | 0.05386 | 12.197 | 2.90 | 4,596,736 |
| | 6 | 64 × 64 × 128 | 4 × 4 × 128 × 64 | 0.175486 | 0.05327 | 11.745 | 2.774 | 8,786,432 |
| | 7 | 128 × 128 × 64 | 4 × 4 × 64 × 64 | 0.349605 | 0.103 | 22.398 | 5.211 | 17,172,736 |
| | | Total | | 1.236147 | 0.37725 | 93.2194 | 20.34 | |
| | | Speedup/ Memory Saved | | | 3.277 | | 4.583 | 35,534,592 |

## 6. Conclusion

This manuscript proposed a novel unified kernel segregation optimization technique for efficiently implementing the transpose convolution layers from the generator part of GANs. The proposed optimized method showed an average speedup of 2.02× (3.89×) using NVIDIA RTX GPU (Intel Xeon CPU) compared to the conventional implementation with the selected datasets. In addition, the proposed method showed memory savings by eliminating an upsampled input feature map. Thus, a unified kernel segregation algorithm did not produce extra elements if the size of the output feature map has odd dimensions. The ablation study showed average speedup above 3× and 4× for GPU and CPU implementation in terms of computation time with the selected GAN models. The proposed method also achieved memory savings above 4.7MB to 35MB from the selected GAN models.